%% file: ClassSR.tex
\documentclass[twocolumn]{aastex63}
\usepackage{amsmath}
\usepackage{gensymb}
\usepackage{booktabs}
\usepackage{xcolor}
\usepackage[ruled, vlined, linesnumbered]{algorithm2e}
\usepackage{algpseudocode}
\usepackage{fontawesome5}
\usepackage{hyperref}
\usepackage{multirow}

\newcommand{\msun}{${\rm M_{\sun}}$}

\def\ltsima{$\; \buildrel < \over \sim \;$}
\def\simlt{\lower.5ex\hbox{\ltsima}}
\def\gtsima{$\; \buildrel > \over \sim \;$}
\def\simgt{\lower.5ex\hbox{\gtsima}}
\def\kpc{{\rm\,kpc}}

\def\msun{{\rm\,M_\odot}}

\def\g{{\rm\,g}}

\def \r {r$^{1/4}$ }

\def\s{\ifmmode \widetilde \else \~\fi}
\def\={\overline}

\def\spose#1{\hbox to 0pt{#1\hss}}

\def\eg{{e.g.,\ }}
\def\ie{{i.e.\ }}
\def\lta{\mathrel{\spose{\lower 3pt\hbox{$\mathchar"218$}}
     \raise 2.0pt\hbox{$\mathchar"13C$}}}
\def\gta{\mathrel{\spose{\lower 3pt\hbox{$\mathchar"218$}}
     \raise 2.0pt\hbox{$\mathchar"13E$}}}
\def\Dt{\spose{\raise 1.5ex\hbox{\hskip3pt$\mathchar"201$}}}    % upper case
\def\dt{\spose{\raise 1.0ex\hbox{\hskip2pt$\mathchar"201$}}}    % lower case

\def\r{${\rm r^{1/4}}$}

\def\dotsfill{\leaders\hbox to 1em{\hss.\hss}\hfill}
\def\sun{\odot}
\def\earth{\oplus}

\def\Gyr{{\rm\,Gyr}}

\def\ltsima{$\; \buildrel < \over \sim \;$}
\def\gtsima{$\; \buildrel > \over \sim \;$}
\def\lsim{\lower.5ex\hbox{\ltsima}}
\def\gsim{\lower.5ex\hbox{\gtsima}}
\def\lapp{\ifmmode\stackrel{<}{_{\sim}}\else$\stackrel{<}{_{\sim}}$\fi}
\def\gapp{\ifmmode\stackrel{>}{_{\sim}}\else$\stackrel{<}{_{\sim}}$\fi}

\definecolor{myred}{HTML}{c00028}
\definecolor{mydarkblue}{HTML}{005353}
\definecolor{myddblue}{HTML}{048080}
\definecolor{myblue}{HTML}{098f94}

\defcitealias{V21}{V21}
\usepackage{subfigure}

\newcommand{\github}[1]{%
   \href{#1}{\textcolor{black}\faGithubSquare}%
}

\submitjournal{ApJL}
\received{December 04 2023}
\revised{June 08 2024}
\accepted{June 16, 2024}

\shorttitle{Class Symbolic Regression}

\shortauthors{Tenachi et al.}
\graphicspath{{./}{figures/}}
\begin{document}

% Paper specific commands
\def\PhySO{{$\Phi$-SO}}
\def\insitu{{\textit{in situ}\ }}
\def\Insitu{{\textit{In situ}\ }}
\def\posthoc{{\textit{post hoc}\ }}
\def\Posthoc{{\textit{Post hoc}\ }}
\def\placeholder{{\square}}

\title{Class Symbolic Regression:\\Gotta Fit 'Em All}
% We change the title it if you want

\correspondingauthor{Wassim Tenachi}
\email{wassim.tenachi@astro.unistra.fr}

\author[0000-0001-8392-3836]{Wassim Tenachi}
\affiliation{Universit\'e de Strasbourg, CNRS, Observatoire astronomique de Strasbourg, UMR 7550, F-67000 Strasbourg, France}

\author[0000-0002-3292-9709]{Rodrigo Ibata}
\affiliation{Universit\'e de Strasbourg, CNRS, Observatoire astronomique de Strasbourg, UMR 7550, F-67000 Strasbourg, France}

\author[0009-0001-0314-7038]{Thibaut L. François}
\affiliation{Universit\'e de Strasbourg, CNRS, Observatoire astronomique de Strasbourg, UMR 7550, F-67000 Strasbourg, France}

\author[0000-0002-8788-8174]{Foivos I. Diakogiannis}
\affiliation{Data61, CSIRO, Kensington, WA 6155, Australia}

% ------------------------------------------------------------------------------------------------------------------------
% ------------------------------------------------------- ABSTRACT -------------------------------------------------------
% ------------------------------------------------------------------------------------------------------------------------

\begin{abstract}
We introduce ``Class Symbolic Regression'' (Class SR) a first framework for automatically finding a single analytical functional form that accurately fits multiple datasets - each realization being governed by its own (possibly) unique set of fitting parameters. This hierarchical framework leverages the common constraint that all the members of a single class of physical phenomena follow a common governing law. Our approach extends the capabilities of our earlier Physical Symbolic Optimization ($\Phi$-SO) framework for Symbolic Regression, which integrates dimensional analysis constraints and deep reinforcement learning for unsupervised symbolic analytical function discovery from data. Additionally, we introduce the first Class SR benchmark, comprising a series of synthetic physical challenges specifically designed to evaluate such algorithms. We demonstrate the efficacy of our novel approach by applying it to these benchmark challenges and showcase its practical utility for astrophysics by successfully extracting an analytic galaxy potential from a set of simulated orbits approximating stellar streams.
\end{abstract}


% ------------------------------------------------------------------------------------------------------------------------
% ----------------------------------------------------- INTRODUCTION -----------------------------------------------------
% ------------------------------------------------------------------------------------------------------------------------

\section{Introduction}
\label{sec:Introduction}

\begin{figure*}
\begin{center}
\includegraphics[angle=0, clip, width=0.8\hsize]{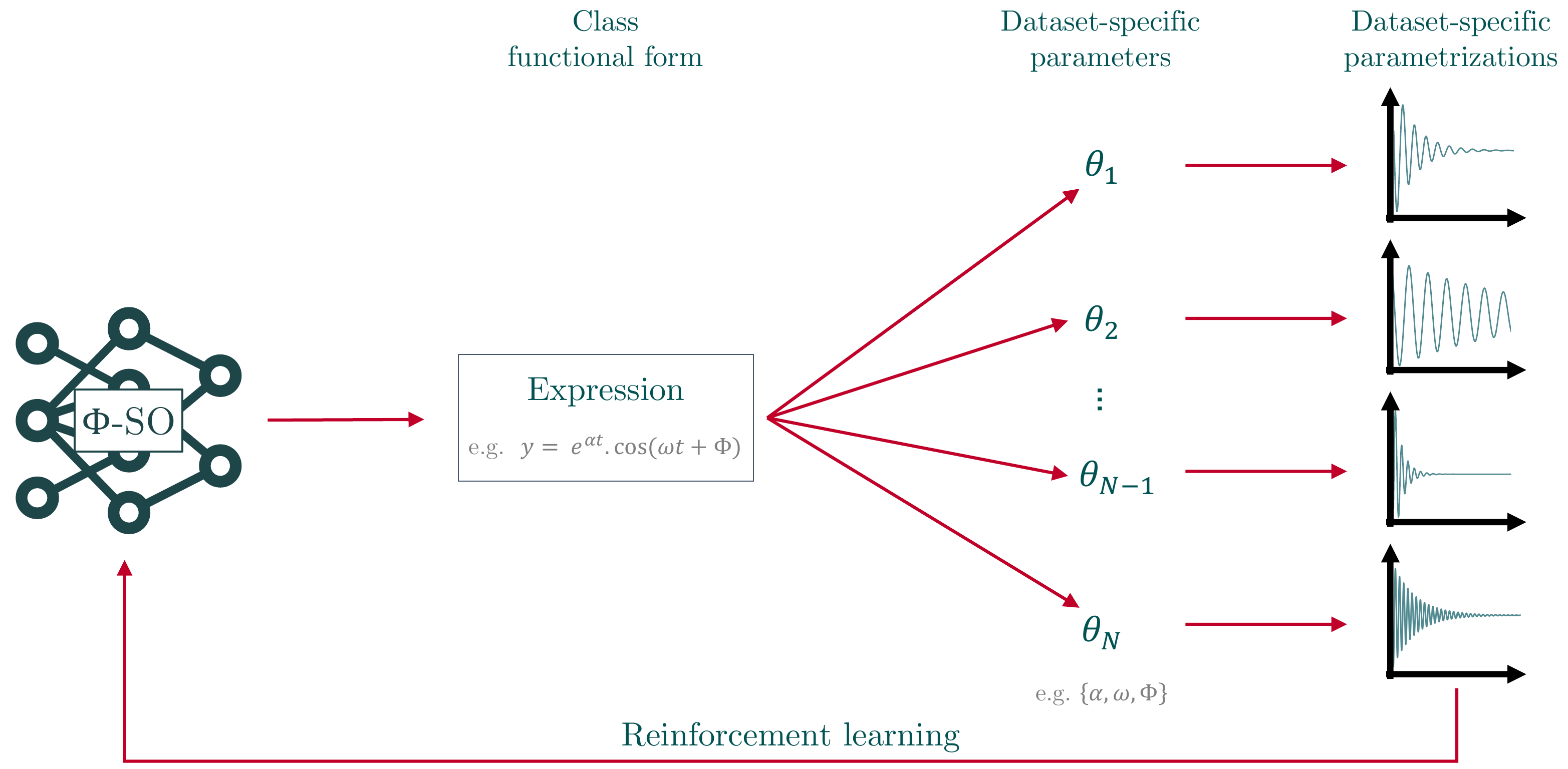}
\end{center}
\caption{Class Symbolic Regression framework sketch: searching for a unique functional form simultaneously fitting multiple datasets. The process starts at the left hand side, a batch of trial class analytical expressions are generated using our \PhySO\ framework \citep{physo_paper}. The free parameters appearing in those expressions are then optimized in a dataset-specific manner \ie allowing each dataset to have its own unique associated values for each parameter. The neural network used to generate the trial expressions is then reinforced based on the fit quality of the trial symbolic functions. This process is repeated until convergence.}
\label{fig:framework}
\end{figure*}

Since the beginning of the scientific revolution, researchers have tried to find repeatable regularities in experiments and observations. Mathematical structures were used in this exploration, and many new ones including functions and differential equations were developed to respond to this need to model nature. Perhaps because of shared symmetries between nature and mathematics, these abstract structures have often been found to work exceedingly well in reproducing and predicting properties of the world, to the point where some have even considered whether the universe is actually mathematical at heart \citep{2008FoPh...38..101T}.

The Symbolic Regression (SR) that the present contribution is concerned with has a long pedigree. Perhaps its most famous application was by Kepler to planetary ephemerides, thereby finding the fitting law that bears his name \citep{1609anov.book.....K}. This empirical law gave the observational basis upon which Newton was able to build the physical theories developed in his Principia Mathematica \citep{1687pnpm.book.....N}. 

Modern SR \citep{EureqaPaper2009, EureqaPaper2011_AFP, OPERON, Exhaustive_SR_wconstraints, Exhaustive_SR, Grammar_prior_wMC_SR, BSR_Bayesian_SR, Divide_and_conquer, GSR, AIFeynman, AIFeynman2, Kamienny_EndToEndSR, Biggio_EndToEndSR, NeSymReS_EndToEndSR, SymFormer_EndToEndSR, Kamienny_learning_mutations, Martius_SymbolsInNN, Sindy_SymbolsInNN, Zheng_SymbolsInNN_DSRbased, EQL_SymbolsInNN, PetersenDSR, uDSR_DSRbased, DGSR, ParFam, OptimalPot, NestedSindy, LLM_SR, Physo_usage_2_turbulence, Physo_usage_3_wave, Physo_usage_4_antenna, SR_interpretability_review,SR_review_Angelis,ParetoFront_SR_RL,DSR_wTransformers_RL,SymQ_RL_SR,AlphaZero_SR_RL,PhySO_usage,SNIP_supervised,Li2024GPT_supervised,MMSR_SR_multimodal_supervised,Botfip_supervised} aims to use the immense computational resources at our disposal to search through possible analytic descriptions in terms of a set of functions and operators (e.g. $x$, $+$, $-$, $\times$, $/$, $\sin$, $\cos$, $\exp$ $\log$, ...) to best fit some numerical dataset $(\mathbf{x}, y)$ we wish to model. Concretely, one seeks some analytic function $f: \mathbb{R}^n \longrightarrow \mathbb{R}$ that fits $y = f(\mathbf{x})$ given those data. It is worth pointing out here that the search space becomes exponentially larger the longer the analytic expression is that we seek to find. Hence the key to SR is to develop efficient schemes to search through the possibilities, and most importantly, to prune out poor choices. 

Our modern computational abilities have allowed us to examine nature in unprecedented quantitative detail, with cameras, spectrographs and other detectors amassing vast quantities of numerical data. It is likely that the clues to next-generation physics and understanding lie therein, and so we are tasked to devise methodologies capable of handling this wealth of information and translating it into coherent, interpretable and intelligible physical models. The promise of SR is that it may allow us in part to answer this need to find accurate and intelligible empirical laws in giant datasets to best capitalize on the community's observational investments.\\

While SR has been extensively applied in scientific research, its focus has largely been on single dataset analysis, overlooking the rich potential in examining multiple datasets linked to a singular physical phenomenon. The present article extends our \PhySO\ framework (presented in \citealt{physo_paper, PhySO_neurips}) further by allowing the search for a functional form that can simultaneously fit several datasets at once, each realization having (possibly) different fitting parameters. This opens up the new possibility of implementing a functional search on the properties of a \emph{class} of objects.
This approach is relevant across various natural sciences, but it particularly shines in astrophysics, where multiple observations of a single phenomenon are often available, providing a rich multi-dataset setup enabling us to devise 'universal' laws that apply to a range of celestial objects of interest.\\

In particular, we apply this new framework to the recovery of a Milky Way-like analytic galactic potential from simulated orbits that can be inferred from stellar streams. Specifically, our approach recovers a single analytical form for the energy of stellar stream members, incorporating a `universal' term that encapsulates the dark matter distribution alongside a nuisance term that accounts for the specifics of individual streams - containing parameters allowed to have object-specific values. Unlike traditional black-box deep learning methods, such as auto-encoders, our method generates a physically meaningful, low-dimensional model in the form of an analytical model.

The layout of the paper is as follows: In Section~\ref{sec:method}, we present the methodology of our approach. Section~\ref{sec:benchmark} details a first benchmark for Class SR, consisting of a series of physics problems designed to assess the performance of Class SR systems; here, we also evaluate our method against these benchmarks. In Section~\ref{sec:streams}, we illustrate the practical application of our method in the more complex scenario of a Milky Way-like potential recovery from orbits. Finally, Section~\ref{sec:discussion}, offers a discussion and a conclusion.
% and realistic scientific

\section{Method}
\label{sec:method}

We build our ``Class Symbolic Regression'' (Class SR) framework on the Physical Symbolic Optimization (\PhySO) framework for SR. This framework combines deep reinforcement learning with \insitu dimensional analysis constraints to construct solutions that avoid physically nonsensical combinations of units. This algorithm currently achieves state-of-the-art performance on physics datasets, and significantly outperforms competitors on the standard Feynman SR benchmark \cite{SRBench} in exact symbolic recovery in the presence of even slight levels of noise (exceeding $0.1\%$). 

Figure \ref{fig:framework} gives an overview of our Class SR framework. Using \PhySO\ we generate a batch of analytical expressions via a recurrent neural network (RNN). In these expressions, class-parameters ($\mathbf{c}$) - which are shared across the entire class and have consistent values across all datasets - can appear alongside realization-specific parameters ($\mathbf{k}$). Subsequently, we optimize the free parameters appearing in each expression $\left(\mathbf{c}, \mathbf{k} \right)$, assigning unique values to realization-specific parameters $\{k_i\}_{i<N_{r}}$ for each of the $N_{r}$ datasets.

This optimization is conducted using the L-BFGS nonlinear optimization routine \cite{LBFGS}. Encoding our mathematical symbols with \texttt{PyTorch} \cite{pytorch}, enables us to use \texttt{PyTorch}'s implementation of the L-BFGS routine, which benefits from \texttt{PyTorch}'s auto-differentiation capabilities to efficiently and simultaneously optimize both class and realization-specific parameters employing a mean squared error (MSE) cost function: $\text{MSE} = \frac{1}{N_{r}. \sum\limits_{i=1}^{N_{r}} N(i) } \sum\limits_{i=1}^{N_{r}} \sum\limits_{j=1}^{N (i)}  \left(y_{ij}-f(\mathbf{c}, \mathbf{k}_{i}, \mathbf{x}_{ij})\right)^2$. Where $\mathbf{x}_{ij}$ are the input variables, $y_{ij}$ are the target values and $N(i)$ is the number of samples which depends on the dataset.

We then use reinforcement learning to update the RNN's parameters following a risk-seeking gradient policy \citep{PetersenDSR}, as detailed in \cite{physo_paper}. This update is based on a reward $R=(1+\text{NRMSE})^{-1}$ that is representative of the fit quality of the trial functional form $f$ across all datasets - evaluated using a normalized root mean squared error (NRMSE): $\text{NRMSE} = \frac{1}{\sigma_y}\sqrt{ \frac{1}{N_{r}. \sum\limits_{i=1}^{N_{r}} N(i) } \sum\limits_{i=1}^{N_{r}} \sum\limits_{j=1}^{N (i)}  \left(y_{ij}-f(\mathbf{c}, \mathbf{k}_{i}, \mathbf{x}_{ij})\right)^2}$, where $\sigma_y$ is the standard deviation of target values evaluated across all datasets. We repeat this process until the RNN converges to a unique high quality expression and its associated parameter values simultaneously fitting all datasets.\\

Furthermore, the sequential nature of expression generation in our \PhySO\ framework enables the incorporation of various priors regarding the resulting expressions as demonstrated in \citep{physo_paper, Bartlett_priors, DSR_priors, DSR_wikipedia_prior}. This allows for customized constraints on the generated expressions such as adherence to the rules of dimensional analysis (which was one of the focal points of our previous study \citealt{physo_paper}) but also simpler priors such as constraints on the number of occurrences of given parameters, the length of the expression and more.\\

\section{Multi-Dataset SR Challenges}
\label{sec:benchmark}

\input{table_benchmark}

% The need for a Class SR benchmark
Despite existing research efforts to establish benchmarks for SR \citep{SRBench, Benchmark_SR_phy, ExpressionSampler_benchmark, EureqaAstroBenchmark}, a benchmark tailored specifically for Class SR has yet to be developed, reflecting the innovative nature of this approach. To address this, we introduce our own Class SR challenges, designed to evaluate a system's ability to analyze multiple datasets. These datasets represent varied observations of a similar phenomenon occurring at different scales but governed by a consistent functional form. Table~\ref{table:benchmark} outlines these challenges, each focusing on accurately recovering the symbolic expression from synthetic datasets having varied scale parameter values. To heighten the challenge, we include multiple scenarios incorporating class parameters that are common to all realizations in addition to other realization-specific parameters.\\

% Benchmarking details
We evaluate our algorithm by randomly sampling $10$ datasets of $10^2$ samples for each of the $8$ challenges described in Table~\ref{table:benchmark} and allowing a maximum of $200,000$ expressions to be explored during each run. In order to ensure robustness, for each challenge, the procedure was repeated $5$ times, each time with a unique random seed, and the recovery rates were subsequently averaged. The whole benchmark tests were conducted across four noise levels: 0\%, 0.1\%, 1\% and 10\% with noise being added individually to each dataset as per the \texttt{SRBench} \citep{SRBench} standardized SR benchmarking protocol : $y_{\text{noise}} = y + \epsilon, \quad \epsilon \sim \mathcal{N}\left(0,\gamma\sqrt{\frac{1}{N}\sum_{i}y_{i}^{2}}\right)$ where $\gamma$ is the level of noise. We conduct runs having access to a single dataset (SR) and having access to all $10$ datasets (Class SR), leading to the total evaluation of $64,000,000$ expressions through $320$ runs.\\

% Hyperparams details
We run our algorithm using the hyper-parameters detailed in \cite{physo_paper}, but with dimensional analysis disabled to ensure a fair comparison with other algorithms (as a consequence the batch size is lowered to $2000$). This adjustment allows future comparisons with our system to be focused solely on the machine learning technique used (here reinforcement learning), rather than the problem simplification achieved through dimensional analysis. We allow the use of the following operations: $\{ +, -, \times, /, 1/\placeholder, \sqrt{\placeholder}, \placeholder^2, -{\placeholder}, \exp, \log, \cos, \sin\}$, a constant equal to one $\{1\}$, two adjustable realization specific free constants $\mathbf{k} = \{k_1, k_2\}$ allowed to have dataset-specific values and one adjustable class free constant $\mathbf{c} = \{c_1\}$. The recovery rate is evaluated by examining each expression in the Pareto front, which contains optimum expressions found in conciseness / accuracy \ie : best fitting expressions at each level of complexity generated by our algorithm. Successful recovery is noted if an expression on the Pareto front is symbolically equivalent to the target expression. Exact symbolic recovery is assessed by formally comparing these expressions with the target expression using the \texttt{SymPy} library for symbolic mathematics \cite{sympy}, following a methodology similar to the one in the \texttt{SRBench} \citep{SRBench}. Specifically, expressions are deemed equivalent if their fraction is symbolically equivalent to $1$ or a constant or if their difference is symbolically equivalent to $0$ or a constant. \\

\begin{figure}
\begin{center}
\includegraphics[angle=0, clip, width=1.\hsize]{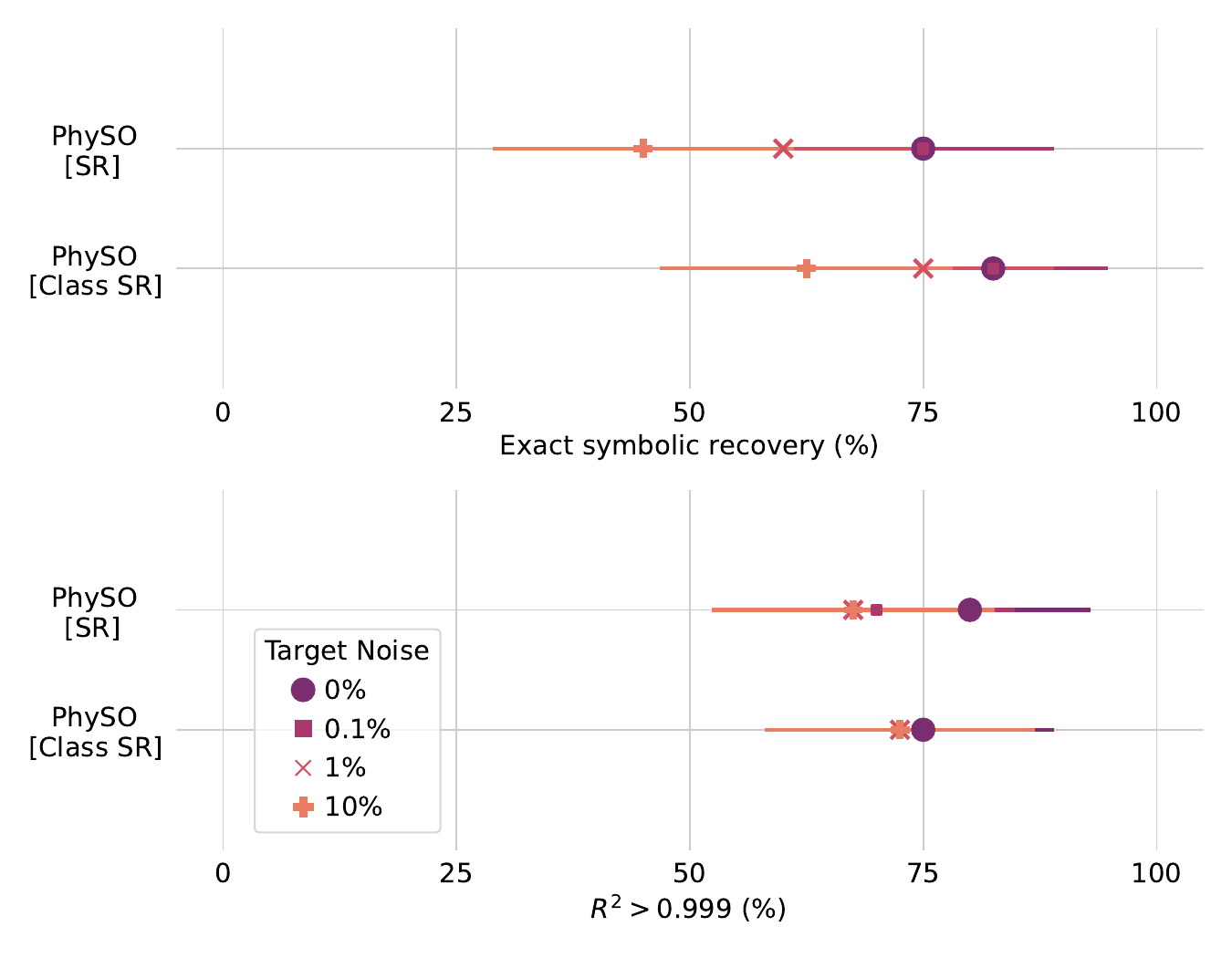}
\end{center}
\caption{Comparison of exact symbolic recovery rates and rate of accurate expressions (having $R^2 > 0.999$) between Class SR and standard SR on our Class SR challenges using an \texttt{SRBench}-style benchmarking pipeline \citep{SRBench}. This figure demonstrates the enhanced effectiveness of Class SR in identifying common underlying functions across multiple datasets with varying scale parameters, resulting in a higher success rate compared to the traditional SR method exploiting only one dataset at a time - especially in the presence of noise.}
\label{fig:class_benchmark}
\end{figure}

% Results overview
Figure~\ref{fig:class_benchmark} presents a comparison of exact symbolic recovery rates  between our Class SR framework and the traditional SR approach under both noiseless and noisy conditions using an \texttt{SRBench}-style benchmarking pipeline, with detailed challenge-by-challenge results published online (see Section \ref{sec:availability}). Our results demonstrate the superiority of Class SR over traditional SR in exact symbolic recovery, particularly in noisy scenarios where noise overfitting is generally an important concern \citep{SRBench}. \\

% Exact symbolic recovery
While one might consider employing traditional SR individually on each dataset and subsequently aggregating the results, this approach would not only be substantially more computationally demanding, but it would also fail to differentiate class constants from realization-specific scale parameters, thus yielding a less interpretable model. Furthermore, our analysis uncovers several instances where traditional SR did not successfully identify the correct expression in any of the $5$ attempts but in which Class SR effectively discovered the correct expressions. This concerns Problem $\#3$ and $\#6$  at 10\% noise level scenarios, as well as Problem $\#5$ across all noise levels. These findings highlight the superior robustness and efficiency of Class SR over traditional methods.\\

% Fit quality
Following the \texttt{SRBench} protocol, we also include, on Figure~\ref{fig:class_benchmark}, the rate of accurate expressions (having $R^2 > 0.999$) with the $R^2$ metric defined as $R^2 = 1 - \frac{\sum_{i=1}^N (y_i - f(\mathbf{x}_i))^2}{\sum_{i=1}^N (y_i - \bar{y})^2}$. We evaluate fit quality by refitting all constants of candidate expressions on newly generated previously unseen test datasets. This approach ensures a fair comparison between Class SR expressions, whose numerical parameters must accommodate multiple observations, and expressions derived from traditional SR, which only fit a single observation. Our results demonstrate that Class SR is not only more efficient at recovering the exact expressions but also more effective at deriving accurate approximations than traditional SR, in scenarios with noise levels exceeding 0.1\%.\\

\section{Recovering an analytic potential form stellar streams}
\label{sec:streams}

\begin{figure*}
\begin{center}
\includegraphics[angle=0, clip, width=1.0\hsize]{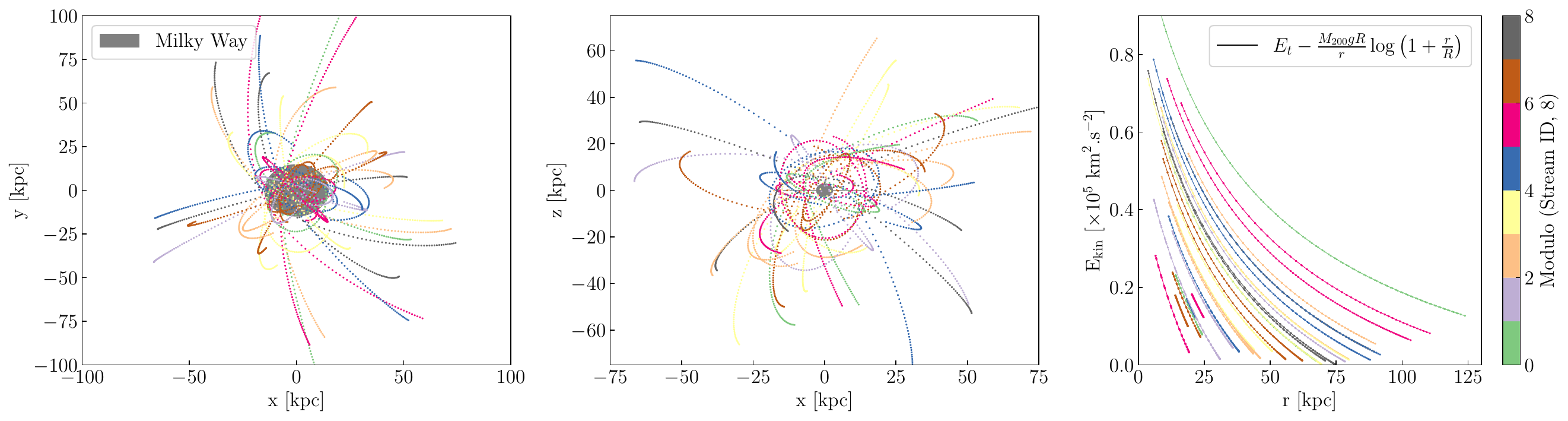}
\end{center}
\caption{Synthetic stellar stream data utilized by our algorithm to recover the galactic potential. The left and middle panels display the spatial positions of stream members relative to the Milky Way, while the right panel illustrates the kinetic energy of these members as a function of their radial distance from the galactic center.}
\label{fig:streams}
\end{figure*}

% Stellar stream problem context
We now turn to an astrophysical application of the algorithm: to try to find the underlying potential of a gravitational system from a set of orbit segments within it. This could be practically applicable for finding an analytic potential model of a galaxy from a set of stellar streams. These linear structures form from the tidal dissolution of globular clusters and dwarf satellite galaxies. When their progenitors are of low mass, the escaping stars have similar energy to the progenitor, and therefore follow a similar orbit. Hence stellar streams approximate orbits in the host galaxy. As has recently been shown by \citet{ChartingMW}, for many real streams one can calculate a ``correction function'' to convert an orbit model into a stream track, and these functions are relatively insensitive to the adopted potential. This procedure can be inverted to give the orbit from the stream.

For this test we imagine having access to full 6-dimensional phase-space measurements of a sample of streams. For each structure $i$, the kinetic energy per unit mass $E_{i, \rm kin}({\bf x})$ is simply:
\begin{equation}
\frac{1}{2}{\bf v}^2 = E_{\rm t}^i - \Phi({\bf x}) \, .
\end{equation}
The total energy per unit mass $E_{\rm t}^i$, which is constant, but different, for each stream, can be considered to be nuisance terms in our search for the underlying potential $\Phi$.\\

% hyperperams details
We run our algorithm with the objective of recovering the analytic form for $E_{i, \rm kin}({\bf x})$. We use the the hyper-parameters detailed in \cite{physo_paper}, allowing the use of the following operations: $\{ +, -, \times, /, 1/\placeholder, \sqrt{\placeholder}, \placeholder^2, -{\placeholder}, \exp, \log, \}$, a constant equal to one $\{1\}$, one adjustable realization specific free constant (having units of energy) and three adjustable class free constants (one having units of energy, one having length units and the other being dimensionless).

% benchmarking details
Again we conduct runs at four noise levels ($0\%$, $0.1\%$, $1\%$ and $10\%$), having access to a single orbit (SR), 25\% of the orbits, 50\% of the orbits and 100\% of the orbits (Class SR), repeating experiments $16$ times with different random seeds and allowing a maximum of $250,000$ expressions to be explored during each run, leading to the total evaluation of $64,000,000$ expressions through $256$ runs.\\

% Stellar stream setup
For the present analysis we generated a sample of artificial orbit data (shown in Figure \ref{fig:streams}) that approximates the sample of 29 thin and long streams studied by \citet{ChartingMW}. To this end we used the present day progenitor positions estimated by \citet{ChartingMW}, and integrated orbits within a universal (NFW) dark matter halo model \citep{1997ApJ...490..493N} that very roughly approximates the large-scale mass distribution in the Milky Way. The adopted potential \citep{2001MNRAS.321..155L} is
\begin{equation}
\Phi_{NFW} = - M_{200} \, g  \log[1+r/R] (R/r) \, ,
\end{equation}
where $M_{200}$ is the virial mass of the halo, $g\equiv (\ln(1+c)-c/(1+c))^{-1}$ is a function of the halo concentration $c$ and $R$ is the scale radius. We chose $M_{200}=10^{12}\msun$, $c=10$ and $R=20.0\kpc$. The orbits consist of 100 phase space points at locations between $\pm1\Gyr$ from the current progenitor location.\\

\begin{figure*}
\begin{center}
\includegraphics[angle=0, clip, width=1.0\hsize]{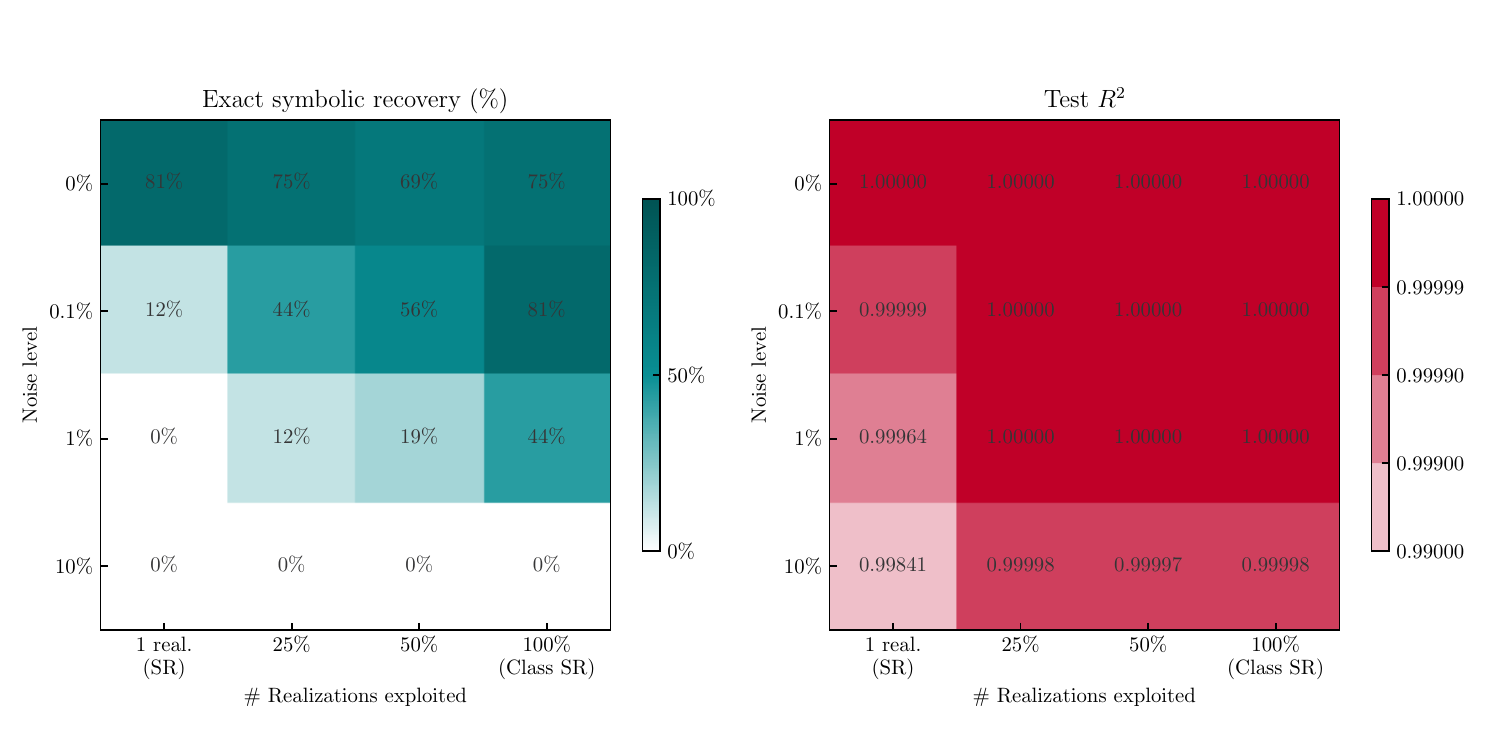}
\end{center}
\caption{This figure presents the exact symbolic recovery rate and median $R^2$ achieved by our Class SR algorithm in the task of recovering an NFW dark matter halo model \citep{1997ApJ...490..493N} from synthetic datasets of stellar stream positions and velocities. The performance metrics are displayed as functions of noise levels and the number of realizations exploited. The edge case, in which a single realization is used, corresponds to the conditions of traditional SR. The results distinctly demonstrate that Class SR substantially outperforms traditional SR, particularly in noisy environments.}
\label{fig:MW_benchmark}
\end{figure*}

% Results overview
Figure \ref{fig:MW_benchmark} presents the results of our analysis in terms of exact symbolic recovery and fit quality, evaluated using the $R^2$ metric. This metric was determined by refitting candidate expressions on noiseless test data and computing the median across various random seeds.

% Exact symbolic recovery and R2
As anticipated, our results underscore that utilizing more realizations during the SR process significantly enhances model accuracy and the likelihood of exact symbolic recovery. This trend is particularly evident as noise levels rise, reinforcing our findings of Section~\ref{sec:benchmark}. Notably, at a $1\%$ noise level, none of the 16 runs that analyzed stellar stream individually succeeded in recovering the correct functional form. In contrast, when all $29$ stellar streams were utilized, the correct functional form was identified nearly half of the time, showcasing the advantages of Class SR under noisy conditions.

% No recovery at 10% noise levels discussion
We observe that the inability of our algorithm to recover the exact symbolic expression in the presence of $10\%$ noise can be attributed to the fact that, under such high noise conditions, the difference in fit quality between the expressions typically identified by our algorithm 
%($E_t -  M_{200} \, g (\frac{r}{R} + \frac{r}{Rc+r})$) 
and the true solution yields only a minimal improvement in terms of reward, $\Delta R \sim 10^{-5}$. This minute improvement, which corresponds to a difference in $R^2$ of approximately ($10^{-6}$), is the sole metric available to guide the algorithm, as it operates on a trial-and-error basis. Unfortunately, such a small difference often remains undetected due to it falling below the tolerance threshold of the free constants optimization procedure. This scenario highlights a known intrinsic limitation of purely empirical SR, where degeneracies in the space of functional forms can go undetected.\\

% Practical case basically works
Excluding scenarios where noise levels render the numerically found expression indistinguishable from the true solution, our Class SR algorithm typically converges toward the correct functional form by exploring under $250,000$ expressions, despite the presence of multiple alternative functional forms that provide a near-perfect fit to individual streams. \PhySO\ identifies an offset parameter specific to each stream (corresponding to $E_{\rm t}^i$) and a functional form parameterized by class-parameters common to all streams corresponding to $\Phi_{NFW}$. These results show that our algorithm can effectively recover a concise intepretable model for a Milky-Way like potential in the form of an analytic expression based solely on stellar positions and velocities without any prior information about the system. 

%\vfill\eject

\section{Discussion and conclusions}
\label{sec:discussion}

We presented a first framework for discovering symbolic analytical functions that simultaneously fit multiple datasets by allowing for (possibly) unique dataset-specific parameter values. This new framework which we dub ``Class Symbolic Regression'' is built upon our earlier \PhySO\ framework which already delivers state-of-the-art performances in symbolic recovery in the presence of noise.

We demonstrated the efficacy of Class SR through simple textbook physics examples which we compiled into a first Class SR benchmark, finding better performance in exact symbolic recovery over traditional SR, especially in noisy situations. Additionally, we applied our method to a more complex astrophysical scenario, successfully rediscovering an NFW galaxy potential model from orbits approximating stellar streams. 

Regular SR, when applied to a single dataset, often risks overfitting to specific characteristics of an observation, such as observational biases or transient events, and noise. In contrast, our Class SR framework should facilitate the finding of \emph{universal} analytical laws that apply to a range of observations within a single class of physical phenomena. This enables our framework to model the underlying physics rather than the specifics of individual observations, with dataset-specific free parameters modeling the unique aspects of each observation. For instance, an application within galactic dynamics that we intend to explore in a future contribution is the analysis of galactic rotation curves. Here, a universal law derived through Class SR could provide insights into the general behavior of dark matter, whereas traditional SR, if applied to a single galaxy, might merely find the specific attributes of that galaxy. \\

It should be noted that while Class SR might superficially resemble regular SR applied to unbalanced datasets with dataset-specific parameters being akin to additional input variables, this comparison is not entirely accurate. In Class SR, these additional degrees of freedom represent unknown values that must be determined, differentiating it as a distinct problem with its own unique challenges.

A persistent issue in SR is model selection as the correct expression can often be overlooked in favor of those that fit better or are less complex (these concerns led to \eg the development of single objective criterion \citealt{Exhaustive_SR}). Our framework, by searching for expressions that fit multiple datasets, effectively utilizes information about the physical phenomena's class structure. This approach significantly mitigates model selection challenges, helping avoid incorrect model choices influenced by dataset-specific peculiarities. In addition, exploiting multiple datasets with regular SR techniques would require fitting the individual datasets independently, and then identifying the solutions in common between the objects, which may not be possible if the measurements are uncertain, would be computationally inefficient and would result in lower performances in exact symbolic recovery and fit quality alike in the presence of noise.\\

Finally, we note that after the first submission of our paper, another Class SR approach built on \texttt{Operon}  \citep{OPERON} - a genetic algorithm approach to SR - was applied to supernovae photometry in \cite{MvSR}.

In future work, we intend to improve on the machine learning aspects of our method to more effectively leverage multiple datasets. As each dataset might distinctly highlight certain symbolic terms or sub-expressions more prominently than others, a promising strategy could be to periodically shift the neural network's training emphasis between datasets. This technique could potentially refine the performance of Class SR by sequentially learning different segments of the expression, rather than attempting to learn the entire expression simultaneously, thereby facilitating the learning process.

\section*{Code availability}
\label{sec:availability}

The documented code for the \PhySO\ algorithm, along with demonstration notebooks, benchmark and results analysis pipelines is accessible on GitHub at \href{https://github.com/WassimTenachi/PhySO}{github.com/WassimTenachi/PhySO} \github{https://github.com/WassimTenachi/PhySO}, complete with comprehensive documentation. A frozen version related to this work on Class Symbolic Regression is released under tag \href{https://github.com/WassimTenachi/PhySO/releases/tag/v1.1.0}{v1.1.0} \github{https://github.com/WassimTenachi/PhySO/releases/tag/v1.1.0}
and deposited on zenodo: \href{https://doi.org/10.5281/zenodo.11663147}{10.5281/zenodo.11663147}.\\

We offer to the community a convenient interface for using our Class SR benchmark, running: \texttt{pb = ClassProblem(i)} will instantiate challenge \texttt{i} $\in \{0, 1, ..., 7\}$ of the Class benchmark presented in Table \ref{table:benchmark}. This interface offers simple ways to generate data points (via \texttt{pb.generate\_data\_points}) and compare a candidate expression to the target (via \texttt{pb.get\_sympy}).\\

In addition, we include challenge-by-challenge and run-by-run performances results tables: see \href{https://github.com/WassimTenachi/PhySO/tree/v1.1.0/benchmarking/ClassBenchmark/results}{\path{PhySO/benchmarking/ClassBenchmark/results}} for results pertaining to the Class SR benchmark and \href{https://github.com/WassimTenachi/PhySO/tree/v1.1.0/demos/class_sr/demo_milky_way_streams/results}{\path{PhySO/demos/demos_class_sr/demo_milky_way_streams/results}} for results pertaining to the stellar stream problem.\\

Finally, for the sake of result reproducibility, we offer a straightforward method to replicate the outcomes presented in Figure \ref{fig:class_benchmark} by simply executing the following command: 
\texttt{python classbench\_run.py --equation i --noise n --n\_reals Nr}. This command will run \texttt{PhySO} on challenge number \texttt{i} $\in \{0, 1, ..., 7\}$ of the Class benchmark presented in Table \ref{table:benchmark}, employing a noise level of \texttt{n} $\in [0,1]$ and exploiting \texttt{Nr} $\in \mathbb{N}$ realizations. We also include the script we used to estimate performances post-run : \texttt{classbench\_results\_analysis.py}

Similarly, we offer a straightforward method to replicate the outcomes presented in Figure \ref{fig:streams} by simply executing the following command: \texttt{python MW\_streams\_run.py --noise n --frac\_real fr}. This command will run \texttt{PhySO} on the stellar stream problem described in Section~\ref{sec:streams}, employing a noise level of \texttt{n} $\in [0,1]$ and exploiting a fraction of \texttt{fr} $\in [0,1]$ realizations. Again, we include our results analysis script: \texttt{MW\_streams\_results\_analysis.py}

\section*{Acknowledgments}
RI acknowledges funding from the European Research Council (ERC) under the European Unions Horizon 2020 research and innovation programme (grant agreement No. 834148).
The authors would like to acknowledge the High Performance Computing Center of the University of Strasbourg for supporting this work by providing scientific support and access to computing resources. Part of the computing resources were funded by the Equipex Equip@Meso project (Programme Investissements d'Avenir) and the CPER Alsacalcul/Big Data.

\bibliography{ClassSR}
\bibliographystyle{aasjournal}

\end{document}

%% file: table_benchmark.tex
\renewcommand{\arraystretch}{1.1}
\begin{table*}[]
\begin{center}
\begin{tabular}{clccc}
\# & Challenge                         & Formula                                             & Variables                                                                  & \begin{tabular}[c]{@{}c@{}}Realization-specific\\ free parameters\end{tabular}                             \\ \hline
1  & Harmonic Oscillator          & $A \cos{\left(\Phi + \omega t \right)}$             & \begin{tabular}[c]{@{}c@{}}$t \in [0.0,9.4]$\\ -\end{tabular}                  & \begin{tabular}[c]{@{}c@{}}$A \in [0.6,1.2]$\\ $\omega \in [0.2,1.5]$\\ $\Phi \in [0.9,1.1]$\end{tabular} \\ \hline
2  & Radioactive Decay            & $n_{0} e^{\frac{-t}{T}}$                            & \begin{tabular}[c]{@{}c@{}}$t \in [0.5,6.0]$\\ -\end{tabular}                  & \begin{tabular}[c]{@{}c@{}}$n_0 \in [0.4,2.0]$\\ $T \in [0.9,1.4]$\end{tabular}                           \\ \hline
3  & Free Fall                    & $\frac{1}{2} 9.81 t^{2} + t v_{0} + z_{0}$          & \begin{tabular}[c]{@{}c@{}}$t \in [0.0,1.0]$\\ -\end{tabular}                  & \begin{tabular}[c]{@{}c@{}}$v_0 \in [-2.0,8.0]$\\ $z_0 \in [-3.0,3.0]$\end{tabular}                       \\ \hline
4  & Damped Harmonic Oscillator A & $e^{- k t} \cos{\left(\Phi + 1.389 t \right)}$      & \begin{tabular}[c]{@{}c@{}}$t \in [0.0,9.4]$\\ -\end{tabular}                  & \begin{tabular}[c]{@{}c@{}}$k \in [0.2,1.0]$\\ $\Phi \in [-0.2,0.3]$\end{tabular}                         \\ \hline
5  & Damped Harmonic Oscillator B & $e^{- 0.345 t} \cos{\left(\Phi + \omega t \right)}$ & \begin{tabular}[c]{@{}c@{}}$t \in [0.0,9.4]$\\ -\end{tabular}                  & \begin{tabular}[c]{@{}c@{}}$\omega \in [0.6,1.4]$\\ $\Phi \in [-0.2,0.3]$\end{tabular}                    \\ \hline
6  & Black Body Photon Count      & $\cfrac{1}{e^{5.9 \nu/T} - 1}$                       & \begin{tabular}[c]{@{}c@{}}$\nu \in [1.0,5.0]$\\ -\end{tabular}                & \begin{tabular}[c]{@{}c@{}}$T \in [1.0,5.0]$\\ -\end{tabular}                                             \\ \hline
7  & Ideal Gas Law                & $\frac{n 8.314 T}{V}$                               & \begin{tabular}[c]{@{}c@{}}$T \in [1.0,5.0]$\\ $V \in [1.0,5.0]$\end{tabular}  & \begin{tabular}[c]{@{}c@{}}$n \in [1.0,5.0]$\\ -\end{tabular}                                             \\ \hline
8  & Free Fall Terminal Velocity  & $\sqrt{\frac{2 m 9.807}{0.47 A \rho}}$              & \begin{tabular}[c]{@{}c@{}}$m \in [1.0,10.0]$\\ $A \in [1.0,5.0]$\end{tabular} & \begin{tabular}[c]{@{}c@{}}$\rho \in [1.0,6.0]$\\ -\end{tabular}                                          \\ \hline
\end{tabular}
\end{center}
\caption{Class Symbolic Regression challenges. Each row details a distinct challenge, with the objective being the exact symbolic recovery of the designated target expression using multiple synthetic datasets. Each dataset being generated using unique realization-specific parameter sets sampled from the given parameter ranges by sampling from the target expression within the given variable ranges.}
\label{table:benchmark}
\end{table*}